# Automatic Generation of Behavioral Test Cases For Natural Language Processing Using Clustering and Prompting

Ying Li[1], Rahul Singh[1], Tarun Joshi[1], Agus Sudjianto[2,3]
[1]Model Risk Management, Corporate Risk, Wells Fargo, USA
[2]H2O.ai, USA    [3]School of Data Science, UNC Charlotte, USA

**ABSTRACT**

Recent work in behavioral testing for natural language processing (NLP) models, such as Checklist, is inspired by related paradigms in software engineering testing. They allow evaluation of general linguistic capabilities and domain understanding, hence can help evaluate conceptual soundness and identify model weaknesses. However, a major challenge is the creation of test cases. The current packages rely on semi-automated approach using manual development which requires domain expertise and can be time consuming. This paper introduces an automated approach to develop test cases by exploiting the power of large language models and statistical techniques. It clusters the text representations to carefully construct meaningful groups and then apply prompting techniques to automatically generate Minimal Functionality Tests (MFT). The well-known Amazon Reviews corpus is used to demonstrate our approach. We analyze the behavioral test profiles across four different classification algorithms and discuss the limitations and strengths of those models.

## 1   Introduction

The advent of deep learning algorithms and transformer architectures has led to significant advances in the performance of natural language processing (NLP) models. However, it is well known that complex models tend to overfit the training datasets and suffer from lack of generalizability. There are many challenges in real-world applications due to the dynamically varying nature of data ([2],[3]) diverse inputs, and sparse training data in some applications. This creates challenges in developing and implementing appropriate test suites to assess model performance. Recently, stimulated by the behavioral testing paradigm in software engineering, researchers have proposed analogous methods for testing NLP models. For example, the paper CheckList ([1]) introduced Minimum Functionality Test (MFT), which are simple test cases designed to test a specific behavior, for example testing the negation, vocabulary, invariance towards the Named Entity recognition (NER) capability of the model. Negation MFTs are simple test cases in which negations are introduced in a single line of text to test whether the model understands negations. These tests are constructed from a manually preset template, and they allow expansion to multiple tests for detecting potential model weakness. However, a major limitation for this approach is that the template needs to be designed manually. The process can be time and resource intensive as they must be developed for domain-specific applications. Further, the quality of the template varies with the subject-matter experts' knowledge, creativity, and language skills. In addition, the overall semantic and syntactic diversity of the tests are restricted due to the fixed format and structure of the base template. It is also possible that the generated test cases might have different data distribution compared to the original dataset.

Automatic Behavioral Test Case Generation Using Clustering & Prompting

This paper proposes a data-based approach to automate the process of creating diverse test suites. It exploits advances in large language models (LLMs), such as Llama ([4]), ChatGPT ([5]), which have emerged as innovative technologies for generating text for a given instruction. There is considerable ongoing research to determine how to utilize these models to increase efficiency. In this paper, we exploit the use of LLMs and statistical techniques such as clustering to automate the development of test cases. We demonstrate in our results that the generated test cases are rich in semantic and syntactic features and have good diversity. Figure 1 provides a high-level summary of the approach, and the steps are as follows:

1) Topic Clustering on input text: the first step is to cluster the input text data into different topic clusters using sentence embeddings [9,17]. This step includes transforming text into embedding vectors using a LLM to capture the semantics, followed by dimensional reduction and clustering. Since, the embeddings are high-dimensional, so as a pre-processing step, we use a dimension-reduction technique to transform the high-dimensional data to lower dimensions while still retaining as much of the information as possible. There are many dimension-reduction techniques in the literature. In our experiments, we have used UMAP ([18]). Then, we apply a topic clustering technique to categorize the dataset (associated embeddings) into meaningful clusters that are diverse across topics. In our experiments, we have found K-means clustering to be effective.
2) Extract Top Representative Documents: Representative documents are then extracted from different topics (i.e., clusters). Hence each topic is represented by a few selective documents. These representative documents are an approximate representation for the entire input dataset.
3) Generate MFT cases by LLM Prompting: A second and a more powerful LLM is used to generate high quality Minimum Functionality Test (MFT) cases based on these representative documents. In this step, the LLM further extracts multiple subtopics within each document and creates test cases within these topics. This allows us to break documents into smaller components and categorize them.
4) Topic Clustering on MFT cases: After generating the MFT cases, we can use LLM model or other techniques, such as business rules, to cluster and categorize the MFT cases into distinct categories.
5) Downstream Model Testing: We prompt users to test the model decisions based on these unit tests.

This process allows us to create small test cases that are diverse, concise and target different properties hidden within input text. LLMs can design test cases from different domains present in the text that let us find the weaknesses of NLP models. New topics/categories will be generated from the automatic MFT test cases. Downstream models will be tested not only on all the MFT test cases but also on the subregions of the MFT test cases through multiple meaningful topics. This allows us to detect model weakness as well as provide explanations for the model.

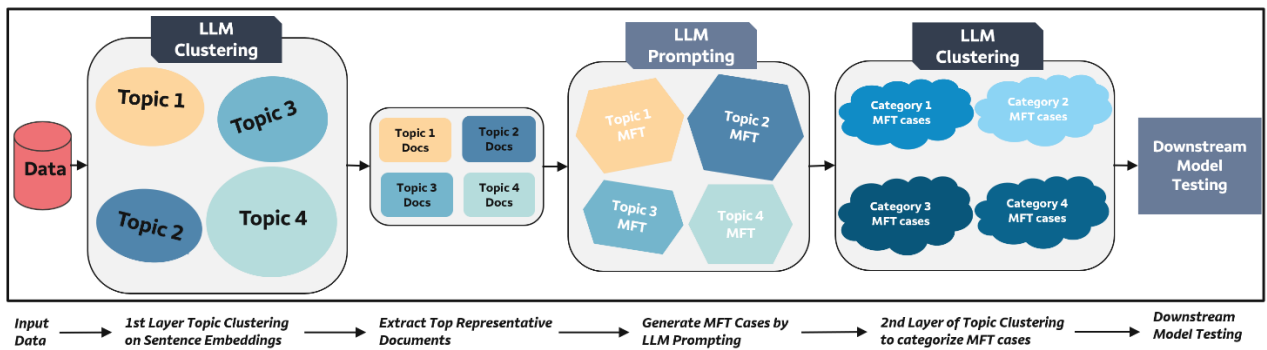

*Figure 1*: Overview of the methodology. First LLM Clustering is applied on the input data to cluster them into several topics and top representative documents are selected from each topic. Diverse Minimum Functionality Test cases are generated from these representative documents by LLM prompting. A second layer of clustering is applied on all the generated MFT cases to get interested categories. Finally, the downstream model is tested on all MFT cases and in subregions.

Figure 2 provides a comparison of our approach with the CheckList approach. The former manually develops MFT and Invariance test (INV) for specific features such as Negation, Named entity recognition (NER). We generate tests for diverse topics automatically from the dataset at hand.

➢ CheckList Paper

➢ Our Target: LLM Topic CheckList

*Figure 2*: Demonstration of our testing methodology with CheckList.
Left: The example shows Negation with MFT and NER with INV that are presented in CheckList Paper.
Right: The example shows test cases in different topics (Customer services, Quality) extracted through LLM.

The rest of the paper develops the details associated with the methodology provided above, applies it to the Amazon review datasets, summarizes the results, and findings. The paper is organized as follows. Section 2 provides information about datasets and models, followed by experiments and results together with practical implementation guidelines. Section 3 discusses model testing for the downstream classification task. Section 4 deals with potential applications, future research directions, and concluding remarks.

## 2 Experiments and Results

### 2.1 Dataset and Model

We utilized the Amazon customer review dataset ([6],[7]) with negative and positive labels. We used a subset extracted from the 6.9 million Amazon dataset in US market with 38 product categories.

*NLP task:* The downstream task is a binary classification problem on this customer review dataset. The original dataset has labels that are one to five stars as ratings. We preprocessed the ratings and relabeled the samples with 1-2 stars as negative reviews and samples with 4-5 stars as positive reviews. Reviews with 3 stars are dropped.

*Text:* The review headline and the review body are concatenated as the text sample for each original record.

*Other data information:* The subset of the customer review dataset includes five categories: mobile apps, books, music, toys, and video.

The dataset is imbalanced consisting of mostly positive reviews with 4 to 5 stars. Down-sampling was implemented on the positive samples so that the final dataset is balanced and has similar numbers of positive and negative labeled samples.

*Train/Validation/Test splits:* We considered the following splits in the data for training the downstream models:

- Training: 32,847 records from 11/11/1995 to 04/13/2014,

- Validation: 7016 records from 04/14/2014 to 12/31/2014, and
- Testing: 7030 records from 01/01/2015 to 08/31/2015.

***Sentence Length***: All three splits contain a negligible percentage of long texts that have over 512 tokens (approximately 3.09% for training and 0.84% for validation, and 0.55% for testing). This makes it ideal for training a transformer classification model, as any excess text would be trimmed to accommodate the limit of 512 tokens imposed by models like BERT-base (or large) ([8]).

***Models:***

1. For sentence embeddings used in the topic clustering, we used gtr-t5-large ([9]) sentence transformer model downloaded from Hugging Face ([10]). It maps sentences & paragraphs to a 768-dimensional dense vector space. The model was trained for sematic search using the encoder from a T5-large model ([11]). The sentence transformer was trained through a loss function to minimize the distance of the embedding vectors between similar sentences while maximizing the distance between dissimilar pairs ([12]).
2. The Llama 2 chat model was utilized for generative tasks in the study. We used a 7 billion parameter chat model ([13],[14]).
3. We did a comparison of four models, including logistic regression, LightGBM ([15]), DistilBERT ([16]) and BERT-base.

## 2.2 Generate Behavioral Test Cases by Topic Clustering and Prompt Engineering

In this subsection, we display the different steps through generating examples which are linguistically diverse but consistent with the original data on topics and data distributions.

### 2.2.1 Topic Clustering and Representative Documents

We started with extracting representative text samples from the original given data. We utilized the BERTopic ([17]) tool for topic clustering and then extracted the representative documents for each cluster (see more details in **Appendix I**). Due to the high dimensional nature of the text embeddings, we first applied dimensional reduction on the embeddings. Then we used clustering algorithms on the low dimensional embeddings to create clusters. Finally, the top keywords/phrases were extracted from each cluster of documents. The final output from topic clustering is cluster-wise topics that were created by concatenating a few top keywords/phrases from each cluster. In the following experiments, we used gtr-t5-large to embed text samples and utilized Uniform manifold approximation projection (UMAP, [18]) for dimensional reduction. The K-means algorithm was used for clustering. **Figure 3** (Middle Panel), which shows a 2D visualization of the 5 clusters that were extracted from the training data.

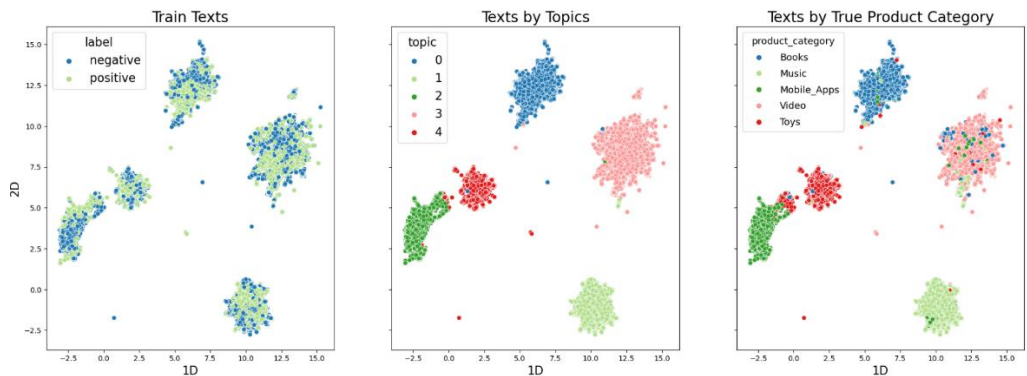

*Figure 3: Two-dimensional text embeddings by gtr-t5-large and UMAP.*
*Left: 2D plot for train text with ground truth labels (positive/negative).*
*Middle: 2D plot for train text by topics.*
*Right: 2D plot for train text by product category.*

For the above data, the K-means clustering algorithm gave five different topics (Middle Panel of **Figure 3**):

- *Topic 0_author_reviews_better_readers,*
- *Topic 1_better_song_songs_bad,*
- *Topic 2_fun_games_addicting_game,*
- *Topic 3_movie_watched_bad_better, and*
- *Topic 4_lego_legos_sets_build.*

*The keywords presented in the topic names clearly reflect the different topics about customer reviews on Books, Music, Game/Mobile Apps, Movie, Toys. A comparison with the right panel of **Figure 3** shows that these identified topics match the ground truth in this dataset.*

*Table 1 provides more details about the documents in each cluster and the number of documents in the "majority" class (positive labels).  **Figure 1**and*

**Table 1** show that the gtr-t5-large embeddings understand the context present in the customer reviews and the proposed approach did well in extracting the product categories of the customer reviews.

| Topic | Count of Samples in this cluster | Majority Product Category in this cluster | Count for Majority Product Category in this cluster | Proportion for Majority Product Category in this cluster | Product Category | Count of Samples in this product category |
|---|---|---|---|---|---|---|
| 0_author_reviews_better_readers (Books) | 7006 | Books | 6806 | 97.15% | Books | 7071 |
| 1_better_song_songs_bad (Music) | 6848 | Music | 6605 | 96.45% | Music | 6933 |
| 2_fun_games_addicting_game (Game/Mobile Apps) | 5527 | Mobile Apps | 4821 | 87.23% | Mobile Apps | 4895 |
| 3_movie_watched_bad_better (Movie) | 9936 | Video | 9552 | 96.14% | Video | 9937 |
| 4_lego_legos_sets_build (Toys) | 3530 | Toys | 3316 | 93.94% | Toys | 4011 |

*Table 1: Number of samples in each topic cluster and the statistics of the majority product category samples in each cluster. The last two columns are counts for the product category in train text for reference.*

Ensuring the diversity of the representative documents is the first step for creating diverse Minimum Functionality Test cases later. To be more specific, "diversity" means limiting text samples with similar context in the representative documents such that the representative documents have a wide coverage for representing the topic cluster samples. For example, "Five stars!", "five stars!!!" and "Five stars! Five stars!" should not be selected to be the representative documents together. For extracting representative documents, we employed the Maximal Marginal Relevance (MMR) algorithm ([19]) with a diversity hyperparameter of 0.5 for removing duplicates (see **Appendix II**). This is an important hyperparameter depending on the diversity of the real data. In the current dataset, there are many reviews that are similar. A potential scenario is that sometimes customers copy previous reviews to create a new review. Hence, the diversity parameter helped in creating diverse samples. **Table 2** shows the effect of samples using the diversity hyperparameter 0.5. When the diversity hyperparameter was set to zero, the model included multiple representative documents with similar content. However, upon implementing the MMR algorithm, a significant increase in diversity in the selected representative documents was observed across multiple trial runs. In **Table 2**, the column before MMR shows multiple representative documents with repetitive expressions. They contain "five stars" and "great album" that are repetitive in multiple documents. After MMR, this repetitive phenomenon was reduced, and diverse representative documents were selected. We manually reviewed representative documents for other topic clusters and the diversity in representative documents was significantly increased after MMR algorithm was used.**Table 2** The MMR algorithm selects documents based on their similarity with the different topic clusters while at the same time increasing the distance with the already selected examples to achieve this.

For the experiments shown in the paper, the top 10 representative documents were extracted from each topic and representative documents were selected using stratified sampling to maintain the same distribution of classes. For example, if one cluster has 40% positive labeled samples and 60% negative labeled samples in total, the extracted top 10 representative documents will have 4 positive samples and 6 negative samples.

| Before: with diversity hyperparameter = 0 | After: with diversity hyperparameter = 0.5 |
|---|---|
| 1. **Five Stars Love his music and great CD**<br>2. **Great album Great album.**<br>3. **Five Stars Great album**<br>4. Noise is ALL I hear. MAKE IT STOP First go read the one and two star reviews....that's all you need to know and you can skip my review. Otherwise here is how I feel.<br />My husband and I fell in love with Night Visions. We took our chances buying that CD after just hearing the mainstream hits and we got it right. I was WAY to quick to buy this album though. I heard I Bet My Life, I fell in love and bought the album the day it released and the first reviews of it here were all 5 star. That is the LAST time I buy a CD without hearing the songs first! This is junk and it is painful and torturous to listen to. Seriously I cant even stand for it to play through. There is so much noise and so little quality and why does every song remind me of another band or something that has already been done? Did anyone listen to this first or were they just pushed and force to release this? I'm so mad and frustrated that I want to just trash it. Someone suggested that this album would grow on me...HAHAHA! NO WAY! I remember when Matchbox 20 released Mad Season in 2000. I was disappointed but as time went on and as I listened more it did indeed grow on me and I fell in love all over. The difference between then and now is the fact that I was ABLE to listen to it over and over. Please please please don't make me listen over and over. I'm really wondering the difference between those who LOVE this and HATE it and my best guess is age. Maybe once you kids hit 30 you'll know good music from bad.<br>5. Defective or Incorrect Country CD This afternoon I got my Wallflower CD which I ordered in early September 2014. I have listed to half the songs on my computer with high quality speakers and I really like this first Diana Krall non Jazz CD. What a change for her. Tomorrow will put the CD into my Blu-ray player to listen to it on my audio system to evaluate the sound quality of the CD and her performance. Why one star? My CD displays the sound track names and album name in Japanese or Korean. It also locked up my Media Player when trying to rip the songs. I tried this CD on two computers, same results. Since I copy my CD's to my computer hard drive and my iPod I have to be able to read song titles and the album name. While ripping to my computer hard drive also got an error message that one of the tracks was defective. It did display in English \\"Jazz & Fusion 2015\\" under the Japanese album name. This is either a defective CD or an Asian version of Wallflower. I have left a long message for Amazon Customer Service to replace my defective CD. I just scanned all the Amazon reviews and there are 4 of us with the same complaints. Looks like poor quality control by the CD manufacturer. Yes, I purchased mine at Amazon. Updated on Feb 10. Amazon CS has been really good and they are replacing the CD and they posted the work around on this issue under a Product Alert. | 1. **Music... I've liked here "different" music a long time- so I bought it...**<br>2. **Great album Great album**<br>3. **Four Stars I really like the songs on this cd.**<br>4. Noise is ALL I hear. MAKE IT STOP First go read the one and two star reviews....that's all you need to know and you can skip my review. Otherwise here is how I feel.<br />My husband and I fell in love with Night Visions. We took our chances buying that CD after just hearing the mainstream hits and we got it right. I was WAY to quick to buy this album though. I heard I Bet My Life, I fell in love and bought the album the day it released and the first reviews of it here were all 5 star. That is the LAST time I buy a CD without hearing the songs first! This is junk and it is painful and torturous to listen to. Seriously I cant even stand for it to play through. There is so much noise and so little quality and why does every song remind me of another band or something that has already been done? Did anyone listen to this first or were they just pushed and force to release this? I'm so mad and frustrated that I want to just trash it. Someone suggested that this album would grow on me...HAHAHA! NO WAY! I remember when Matchbox 20 released Mad Season in 2000. I was disappointed but as time went on and as I listened more it did indeed grow on me and I fell in love all over. The difference between then and now is the fact that I was ABLE to listen to it over and over. Please please please don't make me listen over and over. I'm really wondering the difference between those who LOVE this and HATE it and my best guess is age. Maybe once you kids hit 30 you'll know good music from bad.<br>5. Mostly disappointing reissue The good things: Nice packaging, a real digipack so the discs are protected and not crammed into cardboard sleeves for instant scratching upon removal. Great pics, plenty to read in the booklet. Love that they included the original U.S. vinyl/cassette album cover on the back of the booklet. Nice to finally have the single version of "One of the Living" on CD, which features a different mix from the soundtrack album version.<br /><br />The bad things: As expected, hardly any dynamics remain in the music which is mastered way too loud. The extended rock mix of "Better Be Good to Me" is still the edited early fade-out version which was necessary to fit on the 1997 reissue, losing 40 seconds. It was not at all necessary here as the total time of the bonus disc is only 67 minutes. The first CD appearance of "Keep Your Hands Off My Baby" is also either edited or faded early, losing 15 seconds from the original 3:45. There might be more faults, but this was the point where I got depressed enough to stop listening and write this review. Not recommended except for collectors who just need to own another version of the album. |

*Table 2: Selected representative documents before and after using diversity hyperparameter = 0.5.*

### 2.2.2 Prompt Engineering for Generating Behavioral Test Cases

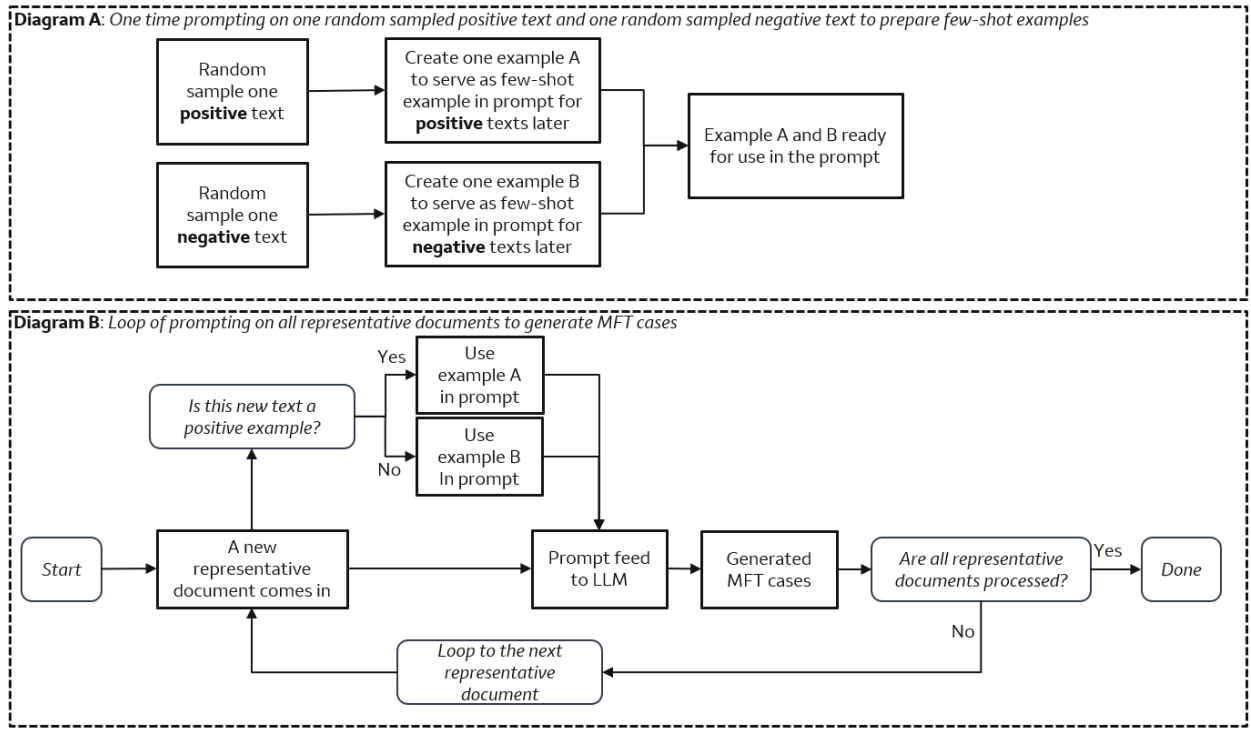

*Figure 4*: Diagrams for creating few-shot examples (Diagram A) and generating MFT cases (Diagram B) through prompt engineering.

We utilized the generation capabilities of a LLM to generate MFT cases by using few-shot prompting as shown in **Figure 4**. The Diagram A in *Figure 4* shows the one-time process for generating few-shot examples by LLMs instead of manually creating these examples. An example is shown in **Figure 5** that shows how to extract test cases and corresponding topics by prompting LLM with some instructions. The Diagram B in **Figure 4** shows the procedure to loop over each representative document for generating MFT cases based on an example input prompt with details shown in *Figure 6* and the example outputs can be found in **Figure 7**.

LLMs can produce high-quality outputs in a specific format with few-shot learning when it is guided by good examples. To achieve optimal results, it is crucial to incorporate meaningful prompts that help the model to understand the type of content to generate. By leveraging LLM's potential, we can generate engaging and informative samples that highlight its capabilities without compromising on accuracy or creativity. In this instance, we use the following prompt to demonstrate how LLM can be employed to construct an illustrative few-shot example as shown in **Figure 5**.

**A RANDOM SAMPLED POSITIVE SAMPLE:**

A masterpiece from a rock legend Since the break up of the Beatles, Paul McCartney for the most part has been a man of singles.  His albums have produced some great singles but much filler acompanied those great songs.  Along comes \\"Chaos and Creation in the Backyard\\".  Hearing this album for the first time is like listening to \\"Pet Sounds\\" for the first time.  The complex sound and maturity of this album is beyond anything McCartney has done in decades.  The flow of the album is amazing.  The last track, \\"Anyway\\", might even be my new favorite solo McCartney song.  I couln't have been happier with this album and I hope any true music lover will feel the same way.  I had to take some time to write the review.  I had to make sure the album didn't get old on me.  It only got better.  So far, this is the best album of the year.

**PROMPT TEMPLATE FOR CREATING THE FEW-SHOT EXAMPLE:**

**prompt** = f"""

<s>[INST]
You are an honest customer. Your customer reviews should be short, simple, clear and real. Always write a customer review with one single sentence.

A Minimum Functionality test (MFT) is simple test cases designed to target a specific behavior. Please extract four minimal functional test samples as customer reviews in one sentence from the following sample and summarize their topics:
{pos_ex_input_text} by filling in the following format:
Test Case 1: [Topic summary]
Customer Review: [Customer review text]

Test Case 2: [Topic summary]
Customer Review: [Customer review text]

Test Case 3: [Topic summary]
Customer Review: [Customer review text]

Test Case 4: [Topic summary]
Customer Review: [Customer review text]
A: Sure! Here are the minimum functionality test (MFT) samples as customer reviews in one sentence for the given sample text and their topics:
[/INST]

"""

**A FEW-SHOT EXAMPLE PRODUCED USING PROMPT TEMPLATE**

Q: Please extract some minimal functional test samples as customer reviews in one sentence from the following sample:
A masterpiece from a rock legend Since the break up of the Beatles, Paul McCartney for the most part has been a man of singles. His albums have produced some great singles but much filler acompanied those great songs. Along comes \\"Chaos and Creation in the Backyard\\". Hearing this album for the first time is like listening to \\"Pet Sounds\\" for the first time. The complex sound and maturity of this album is beyond anything McCartney has done in decades. The flow of the album is amazing. The last track, \\"Anyway\\", might even be my new favorite solo McCartney song. I couln't have been happier with this album and I hope any true music lover will feel the same way. I had to take some time to write the review. I had to make sure the album didn't get old on me. It only got better. So far, this is the best album of the year.
A: Sure! Here are the minimum functionality test (MFT) samples as customer reviews in one sentence for the given sample text:

Test Case 1: Complex Sound and Maturity
Customer Review: This album showcases McCartney's growth as a composer and his ability to create intricate sounds that are both timeless and cutting-edge.

Test Case 2: Flow and Structure
Customer Review: The seamless flow of the album, combined with its thoughtful structure, makes it a cohesive and immersive listening experience that rewards repeated plays.

Test Case 3: Standout Tracks
Customer Review: From the opening notes of "Follow Me," it's clear that this album is something special, with each subsequent track building upon the previous one to create a truly unforgettable listen.

Test Case 4: Emotional Resonance
Customer Review: With its themes of love, loss, and renewal, this album resonates deeply on an emotional level, making it a must-listen for anyone who values authenticity and heartfelt songwriting.

*Figure 5: Prompting for generating one few-shot example.*

Similarly, we can generate a few-shot example based on a randomly sampled negative labeled text.

An important observation in the experiment is that for generating behavioral test cases for one sample, LLM will output better results if the given guided example has the same label as that sample. Otherwise, the output may be influenced by the given guided example in an undesirable manner in practice. In other words, when we provide the prompt to generate behavioral test cases for a positive review, it is essential to use a positive example in the prompt. If we use a negative example in the prompt for this scenario, the sentiment of the generated behavioral test cases for the positive sample may be biased towards the negative example.

While generating these test cases, Llama 2 will not only output several MFT cases but will also output a MFT summary for each generated MFT case such as "Complex Sound and Maturity", "Flow and Structure" etc. as shown in **Figure 5**.

---

**Final Prompt for generating MFT cases for a new sample:**

```
prompt = f"""
<s>[INST]
You are an honest customer. Your customer reviews should be short, simple, clear and real. Always write a customer review with one single sentence. Always answer the question following the given example format.

A Minimum Functionality test (MFT) is simple test cases designed to target a specific behavior. For example,
{prompt_example}

Q: Please extract some minimal functional test samples as customer reviews in one sentence from the following sample:
{input_text}
A: Sure! Here are the minimum functionality test (MFT) samples as customer reviews in one sentence for the given sample text:
[/INST]
"""
```

---

*Figure 6: Prompt for generating MFT cases. {prompt_example} is a few shots example created above. {input_text} is the current text sample that we want to generate MFT cases from.*

When a text comes in for generating MFT cases, we feed few shots of guided examples with the same label to the prompt as shown in **Figure 6**.

We iterate through this process on all selected representative documents across all topic clusters. The LLM will follow the output format guided by the given examples using few-shot learning so that it enabled us to automatically extract all the outputs into a structured table easily. The generated MFTs have substantial amount of semantic richness as well as proper sentence structure in terms of its overall performance. Two instances of MFT cases resulting directly from the provided reviews are displayed graphically in **Figure 7**. In addition, there are several illustrations in the next table (**Table 3**) of what MFT cases can be created from reviews originating from distinct product categories.

---

**Text:** It is tiny! I know that legos are expensive, but my gosh! This box is so small!! My 7 year old got it for Christmas and has no interest in it! It was so expensive for what I got! I would not recommend buying this at full price at all. We like legos, just not that much I guess.
text true label 0

**Output MFT cases:**
Test Case 1: Disappointment with size
Customer Review: Tiny box for the price! Not worth it.

Test Case 2: Lack of interest from children

Customer Review: Kids aren't impressed with this set. Too small.

Test Case 3: Overpricing of product

Customer Review: Expensive for what you get. Not worth full price.

**Text:** so, addictive on kindle fire This game is way more fun than several other free games ive tried. This is now in my top 5 favorite games! I used it on my kindle fire and love it.
text true label 1

**Output MFT cases:**

> Test Case 1: Addictiveness
> Customer Review: This game is so engrossing that I found myself unable to stop playing it, even after multiple sessions.
>
> Test Case 2: Favorite Game
> Customer Review: Now, this is what I call a truly enjoyable game - it's moved into my top five favorites on Kindle Fire, and I can't wait to play it again.
>
> Test Case 3: User Experience
> Customer Review: The interface is intuitive, and the gameplay is smooth and responsive, making for a delightful user experience that keeps me coming back for more.
>
> Test Case 4: Replay Value
> Customer Review: Despite having played through the entire game multiple times, I still find myself wanting to come back and try different strategies or challenge myself in various ways - the replay value is simply incredible.

*Figure 7: Examples of MFT cases generation.*

| MFT Summary | MFT Case: Review Text | Source Review Product Category |
|---|---|---|
| Lack of character development | No character development whatsoever; it feels like a bunch of random strangers were thrown together without any context or backstory. | Books |
| Disappointment with the album's overall impact | I was really looking forward to this album, but it ended up being a huge disappointment. There's nothing here that stands out or grabs my attention. | Music |
| Buggy Gameplay | Keeps kicking me off after initial enjoyment. 😒 | Mobile Apps |
| Flaws and Disappointments | Unfortunately, the movie's poor dialogue and underwhelming performance from certain actors detract from the overall experience, particularly in the scenes featuring Jake Lloyd and the two-headed announcer. | video |
| Enjoyable Build Experience | The building process was surprisingly enjoyable, with clear and concise instructions that made it easy to navigate despite the numerous components involved. | Toys |

*Table 3: Examples of generated MFT cases for reviews in different product categories.*
*The column "MFT case: Review Text" is the generated MFT case by llama2.*
*The column "MFT summary" is the generated summarized topic for the MFT case output by llama2.*
*The column "Source Review Product Category" is the product category of the original customer review in the prompt from which llama2 generate the MFT case.*

## 2.3 Multiple Versions of MFT cases

In this paper, we created about four MFT cases for each representative document and in total we created 50 representative documents (10 top representative documents are extracted in each cluster and we have five clusters). After completing the entire procedure, we have around 200 MFT cases after deduplication. For robustness, we run this process three times with different random seeds so that we have three sets of MFT test cases with 200 texts. To enhance the generalizable testing of a particular model in production, we generated MFT instances using the training data, while simultaneously creating MFT instances utilizing the test data. Furthermore, to extend the generalization testing, we asked LLM to paraphrase the MFTs using prompt engineering. We assume the paraphrased MFTs will have the same label as the original MFT. Here is an example output with five alternate versions of the same MFT (**Table 4**).

| MFT summary | Original MFT case | Output Five Rephrased Versions by LLM |
|---|---|---|
| Poorly developed plot or characters | Disappointing read with too much monotony. | 1. "Underwhelming experience with too much repetition."<br>2. "Disappointing read with too much dullness."<br>3. "Monotonous content without any excitement."<br>4. "A lackluster reading experience with too much redundancy."<br>5. "Uninspired writing with too much repetitive material." |

*Table 4: Example of generated rephrased versions for one MFT case by Llama2.*

### 2.4 Quality Control of the label and generated MFT cases

#### 2.4.1 Minimum Functionality

Although the test cases generated by the MFT method are based on representative documents, will they be straightforward enough for the model to comprehend and categorize? The visualization of the 2D reduced embedding of these MFT cases from the gtr-t5- large sentence transformer suggests they are fundamental and are easily distinguishable by the model as shown in **Figure 8**. As we observe, the embeddings of gtr-t5-large have distinct boundaries between positive (green points) and adverse (blue points) evaluations for MFT instances in **Figure 8**On the other hand, the 2D representation of the original data is a mixture of positive and negative sentiments and the sentiment boundaries are unclear as shown in the left sidebar of **Figure 3**. This comparison shows the MFTs have simple and clear sentiments that are even understood by a general pretrained sentence transformer model while the original texts have more complicated and confusing sentiments to confuse the sentence transformer model. This demonstrates that the MFT are straightforward text examples representing the original data distribution, and we expect that a good model should understand them and give correct results for these test cases.

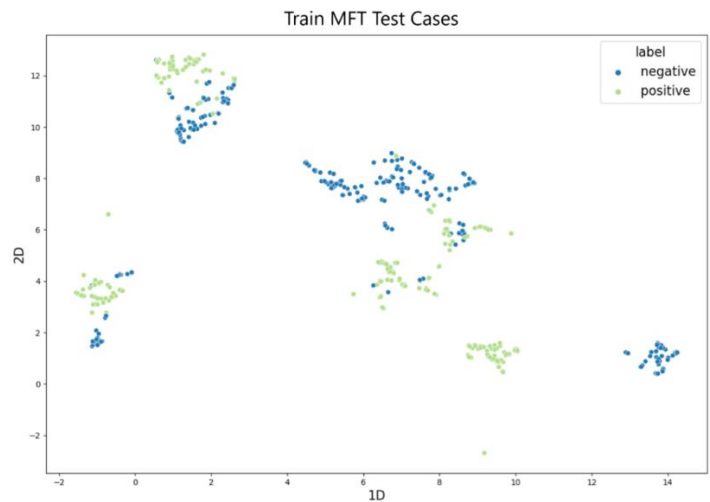

*Figure 8: 2D plot for training MFT cases text embeddings and their labels.*

#### 2.4.2 Label Quality Control

A fast and efficient method for labeling MFT cases involves using the labels of the corresponding original representative documents used to generate the MFT. This approach assumes that the MFTs are labeled with the same label as the original document since they are created from it and retain the same meaning. Nevertheless, this conjecture is approximate may and, in some cases, not be correct because the dataset consists of reviews displaying diverse sentiments. Some evaluations express both favorable and unfavorable perspectives regarding the product yet are classified into just "positive" or "negative". Furthermore, when creating MFT instances, they might be established upon either the positive subsections or the negative subsections, which could result in erroneous label-preserving. For example, consider the review, "The toy has good quality, but the price is too high" with a "negative" label may generate the following two MFT cases 1. "The quality of the toy is pretty good!" and 2. "This toy is too pricy. It isn't worth it.". Obviously the first one should be labelled as a positive review instead of sharing the negative label with the original review text. Hence, it should be safer to do some label quality control before using these unit test cases for testing any models to make sure the data quality of the MFT cases in advance. To expedite the manual labeling process, we asked Llama2 to label it first followed by a manual check (*Figure 9*).

---

**Review (input_text):** "I don't like the product"

**Prompt:**

```
prompt = """
<s>[INST]
You are a reliable annotator. Always answer the question following the given example format.

Q: Please help me to label the sentiment of following text into 1.Positive, 2.Negative or 3.Hard to Decide: {input_text} by filling in the following format.
A: Of course! I'd be happy to help you with that. Here's my response:
Label: [Fill label here]
Reason: [Fill reason here]
[/INST]
"""
```

**Output:**

Of course! I'd be happy to help you with that. Here's my response:
Label: Negative
Reason: The sentence expresses dislike towards a product, indicating a negative sentiment.

---

*Figure 99: Prompting for labeling new MFT cases by Llama2.*

Manual checking revealed the high accuracy of labelling by Llama2 as we found only 10 samples that are mislabeled. Most of these cases have multiple emotions, both positive and negative, and hence Llama2 struggled to label it correctly. We removed these "hard" samples and retained only the unambiguous cases.

### 2.4.3 MFT Topic Clusters

It is possible that the MFT cases might have different topics compared to the original data. We can see this through four clusters for the MFT cases in **Figure 1010** when we compare it to the clusters of original data in **Figure 3** where there are five clusters. We did another layer of topic clustering on embedding vectors for MFT texts using gtr-t5-large sentence embeddings with BERTopic package. After this second round of clustering, we got four new topics instead of the original five product categories that have more emphasis in review text contents as:

- *"Topic 0: Books/Movies Contents",*
- *"Topic 1: Toys Quality,*
- *"Topic 2. Music",*
- *"Topic 3: Mobile Apps/Toys User Experience".*

After this step, the MFTs are able to reflect the same product categories information as before but it treats books and movie reviews in one large group as the customer reviews in these two groups talk about related topics like plots, character development etc. Furthermore, the clustering now combined part of the toys review with mobile apps reviews as they describe user experience with similar text.

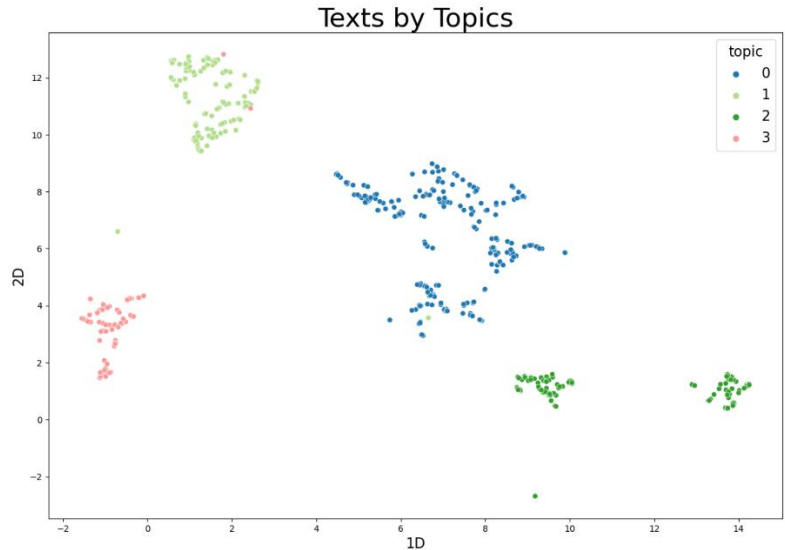

*Figure 1010: Topic clustering on MFT cases embedding vectors.*

This second layer of topic clustering can help us to test and monitor the model performance in distinct groups of texts from a distinct perspective and at the same time help us to detect potential model weakness and explain the model in the subregions of topics.

This is not the only approach to identify secondary topics. In fact, there are multiple ways to get additional topics so that we can test a model from different perspectives. For example, one can do clustering on the MFT Summary (See "MFT Summary" column in **Table 3**) generated by Llama2 when it generates the MFTs. Another way is to group these MFT tests manually with domain specific business acumen and guidelines. As an illustration, these MFT instances may be grouped into sets purely based on their contents instead of product category. For example, after analyzing the MFT context by some business rules with respect to domain specific guidelines, they can be grouped into data quality, customer services, return policy and shipping speed etc. as both book reviews and toy reviews may talk about the similar contents related to these new topics regardless of product category.

## 3 Model Testing

The MFT test cases are now ready to use. In this section, we explore the initial application of testing downstream classification models using these high quality MFT cases.

### 3.1 Testing Scenarios

The performance is reported on different versions of the MFTs, including:

1. Performance on train MFT cases, namely **Train MFT 1**, **Train MFT 2** and **Train MFT 3** datasets generated from training data, respectively.
2. Performance on **Train MFT (Original)**: all train MFT cases by combining Train MFT 1, Train MFT 2, Train MFT 3 datasets together and then apply deduplication.
3. Performance on **Train MFT (Extended)**: all train MFT cases, i.e., Train MFT (Original), and all their five different paraphrased versions followed by deduplication.

The data sizes for different versions of MFT cases are listed in **Table 5** below. The same testing was applied to test MFT cases, i.e., MFT cases generated from test data, for checking model generalization ability further to have better insights about whether the model understands the linguistic context well or not.

| Dataset from Train | Data Size | Dataset from Test | Data Size |
|---|---|---|---|
| Train MFT 1 | 196 | Test MFT 1 | 200 |
| Train MFT 2 | 195 | Test MFT 2 | 200 |
| Train MFT 3 | 192 | Test MFT 3 | 196 |
| Train MFT (Original) | 558 | Test MFT (Original) | 474 |
| Train MFT (Extended) | 3304 | Test MFT (Extended) | 2825 |

*Table 5: Sizes of different versions of MFT cases datasets.*

## 3.2 Test Results

### 3.2.1 Performance on Individual MFT Dataset

| Dataset / Model | Train | Train MFT 1 | Train MFT 2 | Train MFT 3 | Test | Test MFT 1 | Test MFT 2 | Test MFT 3 |
|---|---|---|---|---|---|---|---|---|
| **TFIDF logistic** | **98.77%** | 82.20% | 89.74% | 92.70% | 92.83% | 86.00% | 89.50% | 83.67% |
| **TFIDF LightGBM** | 91.38% | 81.12% | 80.51% | 89.58% | 91.66% | 78.00% | 82.00% | 80.61% |
| **DistilBERT** | 98.74% | **98.96%** | 93.85% | 97.92% | 95.18% | 94.50% | 97.00% | **94.28%** |
| **BERT** | 98.60% | 97.45% | **95.38%** | 98.96% | 95.83% | 96.50% | 98.00% | 93.88% |

*Table 6: Model performance (accuracy) on multiple versions of MFT datasets for the binary classification task.*

To start with simple and basic MFT datasets, we generated ~200 MFT cases for each round and created three versions using three different random seeds. **Table** 6 shows that the BERT model outperforms other models on most of the MFT test datasets except two instances where DistilBERT is slightly better. The performances of logistic and LightGBM models vary more across the three different MFT cases compared to the two transformer-based models. When we iterated this process to create different versions of MFT dataset, the performance on these different versions can be used to calculate the standard deviation of the performance scores and can be a measure of model stability and robustness.

### 3.2.2 Performance on Combined and Extended Train/Test MFT Datasets

To enhance the reliability of the model's accuracy, we tested it on a comprehensive set of MFT cases (approximately 600 instances). The accuracy scores for every model are included in **Table 7**, and they indicate a consistent level of accuracy between the train and test splits of original data. Specifically, we noticed that both the logistic and LightGBM models exhibited a significant decrease in accuracy when transitioning from train-test split to the train-test MFT cases despite having accuracies greater than 90% on the test split of dataset. This reveals the importance of evaluating a model's capacity to adapt outside of the train and test datasets using MFT cases since relying solely on test dataset might result in ignoring potential difficulties.

On the other hand, DistilBERT demonstrated a minimal drop in functionality (less than 1%) shifting from train–test split accuracy to MFT cases. In contrast, BERT was more stable across various datasets.

We further extend generalization testing by evaluating on a broader collection of MFT instances, including the standard MFT samples and five distinctly paraphrased variations generated using Llama2. Our findings indicate that there was a noticeable decline in the performance of logistic and LightGBM models when moving from MFT set derived from the training data (MFT(Original)) to a larger set test enhanced with MFT cases by rephrasing. In contrast,

DistilBERT and BERT showed steady performance across both evaluation sets. Overall, the MFT cases boost our ability to test the model weakness and its generalization ability.

| Dataset / Model | Train | Train MFT (Original) | Train MFT (Extended) | Test | Test MFT (Original) | Test MFT (Extended) |
|---|---|---|---|---|---|---|
| **TFIDF logistic** | **98.77%** | 88.85% | 86.14% | 92.83% | 86.44% | 82.44% |
| **TFIDF LightGBM** | 91.38% | 83.45% | 76.45% | 91.66% | 80.30% | 74.73% |
| **DistilBERT** | 98.74% | 96.58% | 96.43% | 95.18% | 94.28% | 94.09% |
| **BERT** | 98.60% | **97.12%** | **97.09%** | **95.83%** | **95.97%** | **95.79%** |

*Table 7*: **Model performance (accuracy)** on 1. **Original** combined MFT dataset (MFT dataset 1 + MFT dataset 2 + MFT dataset 3) and 2. **Extended** MFT dataset (the original combined MFT test cases and their five rephrased versions) for the binary classification task.

### 3.2.3 Performance on Different MFT Topic Clusters

This section reports model performance across different MFT topic clusters. This allows us to analyze potential model weaknesses on different topics and provide model weakness explainability in terms of topics We see from **Table 8 that**, while LightGBM performs poorly overall, it has better performance for Toys Quality compared to others. Other models show similar performance across the four different MFT topics. To cater to business needs, a separate set of MFT topics can always be defined to test the model from a new perspective.

While LightGBM has the worst performance compared to other models and this is visible from its test set performance (**Table 7**), this analysis provides additional evidence. Here we are mainly demonstrating the idea of finding specific topic of MFTs where the model is weak at and better at. This is useful for finding subregions where the model weakness is.

| MFT Topic / Model | Books/Movie Contents | Toys Quality | Music | Mobile Apps/Toys User Experience |
|---|---|---|---|---|
| **TFIDF logistic** | 84.59% | 88.43% | 86.74% | 86.21% |
| **TFIDF LightGBM** | 74.01% | 84.14% | 73.48% | 75.31% |
| **DistilBERT** | 96.59% | 96.36% | 96.57% | 95.88% |
| **BERT** | **97.09%** | **97.53%** | **97.04%** | **96.50%** |

*Table 8*: *Model performance (accuracy) on sub-dataset of MFT cases with different topics.*

## 4 Discussion and Concluding Remarks

In this paper, we presented a new technique to automatically generate test cases for assessing NLP models. It allows for a comprehensive evaluation of models across various domains and topics that may also serve as a method to explain these models. By doing so, we can gain a better understanding of their capabilities and limitations. For the Amazon reviews dataset, the method was effective in identifying diverse clusters and creating relevant test cases for each cluster.

The main purpose of this paper is to demonstrate the new idea for using clustering and prompt engineering with the help of LLMs to diagnose downstream models through the results on US Amazon Review dataset. This is an ongoing work. We are experimenting on additional datasets and will update the results.

The method has potential applications in other areas.

Automatic Behavioral Test Case Generation Using Clustering & Prompting**Model Robustness:** We have demonstrated that the MFT cases can help us uncover the model's generalization weakness that is hidden in the test accuracy, which will help a lot in making a model selection decision. Beyond that, model robustness can also be monitored and detected among different topics by generating MFT cases with different topics.

**Model Explainability:** The study aims to investigate the model's performance and behavior through various analyses of its responses to MFT test cases with diverse subjects. This comprehensive evaluation will provide valuable insights into the model's strengths and weaknesses across different domains of the data, offering an unprecedented level of interpretability. By examining the model's output for each topic, we can identify areas where the model excels or struggles, allowing for targeted improvements to enhance its overall performance. Furthermore, this study demonstrates the importance of adopting a multi-faceted approach to explainability, recognizing that no single method can fully capture the intricacies of a complex AI system. By integrating multiple techniques, including topic modeling and feature importance analysis, we can gain a more complete understanding of the model's decision-making process, leading to better transparency and accountability in AI development.

In addition, the LLM models reduce time and labor but at the same time generate diverse linguistic tests.

Extending this idea, the next steps will be focused on two main aspects: generating difficult MFT tests and systematically generating them for challenging the model. The former includes but not limited to a more automatic process for checking the label quality automatically (for example, we may ask multiple LLMs to mimic multiple annotators to label the text and cross validate each other's labeling), iterative prompting process to efficiently produce high quality test cases, automated post-evaluation of the data quality control for the MFT cases (such as automatic scores like faithfulness etc. to filter out bad text and keep the high quality ones). Another direction would be to expand it to utilize the current workflow to challenge distinct types of LLM and natural language generation (NLG) tasks.

## Acknowledgements

The authors thank Vijayan N. Nair for supervision, discussion, comments and support for this research and manuscript.

## Appendix

### I. Representative Documents

Instead of designing templates manually, we start with extracting representative text samples from the original given data.

BERTopic is a useful and popular tool for topic clustering on NLP text data. The method by default is following the procedure of first applying UMAP dimension reduction on the text embeddings, then use HDBSCAN clustering algorithms to create clusters for the given data, followed by a Class-based TF-IDF(c-TF-IDF) and KeyBERTInspired algorithm to calculate and finetune the top keywords in each cluster. The final output would be topics which are created by concatenating a few top keywords in each cluster, respectively.

In the KeyBERTInspired algorithm, there is an intermediate step for extracting the representative documents. It is later used for re-ranking the raw topic keywords from c-TF-IDF to downgrade less important keywords and de-noise the raw keywords list.

In detail, on one hand, the algorithm first extracts top words per topic based on their c-TF-IDF scores, which is calculated by the following equation (1).

The c-TF-IDF for a term $x$ in class $c$ is defined as

$$W_{x,\,c} = \left\|tf_{x,\,c}\right\| \times \log\left(1 + \frac{A}{f_x}\right) \tag{1}$$

where $tf_{x,\,c}$ is the frequency of word $x$ in class $c$, $f_x$ is the frequency of word $x$ across all classes and $A$ is the average number of words per class.

On the other hand, the algorithm samples a few candidate documents (500 by default) per cluster. Then, the top n (5 by default) representative documents are extracted by calculating the c-TF-IDF representation for the candidate documents and finding which are closest to the topic c-TF-IDF representation through cosine similarity.

The raw top n words per topic based on the keywords is further finetuned by comparing with the representative documents. The algorithm embeds the candidate keywords, and it also embeds the representative documents followed by averaging. Then it compares the embedded keywords with the embedded documents through cosine similarity scores and re-orders the raw keyword list into the finetuned keyword list by these similarity scores in descending order.

Here, we utilized the BERTopic tool to extract the representative documents for each cluster with some adaptation considering the ground truth label distribution of each cluster to avoid always picking the positive or negative documents.

## II. Maximal Marginal Relevance (MMR)

When the BERTopic algorithm calculates the representative documents in the default hyperparameter setting, it doesn't consider the similarity among the representative documents. For example, when it selects "Five stars!!!" as the next representative document, it will not take this because it is too similar to the previous selected representative document "Five stars". To increase the diversity of the representative documents for a better semantic coverage, the Maximal Marginal Relevance is employed by setting the diversity hyperparameter in BERTopic. This algorithm is trying to select a representative document which is closer to the topic c-TF-IDF at the same time more diverse from other representative documents through the optimization function (2)

$$MMR = \arg\max_{D_i \in R \setminus S}[(1-\lambda)Sim_1(D_i, Q) - \lambda \max_{D_j \in S} Sim_2(D_i, D_j)] \tag{2}$$

where $\lambda$ is the diversity hyperparameter, $R$ are all sampled candidate documents, $S$ are the current set of selected representative documents, $Q$ is the topic c-TF-IDF vector, $D_i$ is the c-TF-IDF vector for a candidate document $i$, $D_j$ is the c-TF-IDF vector for an already selected representative document $j$. With a larger $\lambda$ setting, MMR will penalize more on the similarity between a candidate document and current representative documents when it tries to pick the next top representative document.